\documentclass{article}

\usepackage{PRIMEarxiv}

\usepackage[utf8]{inputenc} 
\usepackage[T1]{fontenc}    
\usepackage{hyperref}       
\usepackage{url}            
\usepackage{booktabs}       
\usepackage{amsfonts}       
\usepackage{nicefrac}       
\usepackage{microtype}      
\usepackage{lipsum}
\usepackage{fancyhdr}       
\usepackage{graphicx}       
\usepackage{amsmath}
\usepackage{pgf-pie}
\usepackage{float}
\usepackage{caption}
\usepackage{svg}
\usepackage{multirow}
\usepackage{algorithm}
\usepackage{makecell}
\usepackage{algpseudocode}
\graphicspath{{images/}}     

\makeatletter
\def\blfootnote{\xdef\@thefnmark{}\@footnotetext}
\makeatother

\usepackage{xcolor}

\title{

\textbf{Efficient RLHF: Reducing the Memory Usage of PPO}

}

\author{
  Michael Santacroce, Yadong Lu, Han Yu, Yuanzhi Li, Yelong Shen \\
  Microsoft \\
  \texttt{\{misantac,yadonglu,hanyu,yuanzhili,yelong.shen\}@microsoft.com} \\
}

\begin{document}
\maketitle

\begin{abstract}

Reinforcement Learning with Human Feedback (RLHF) has revolutionized language modeling by aligning models with human preferences. However, the RL stage, Proximal Policy Optimization (PPO), requires over 3x the memory of Supervised Fine-Tuning (SFT), making it infeasible to use for most practitioners. To address this issue, we present a comprehensive analysis the memory usage, performance, and training time of memory-savings techniques for PPO. We introduce Hydra-RLHF by first integrating the SFT and Reward models and then dynamically turning LoRA "off" during training. Our experiments show: 1. Using LoRA during PPO reduces its memory usage to be smaller than SFT while improving alignment across four public benchmarks, and 2. Hydra-PPO reduces the latency per sample of LoRA-PPO by up to 65\% while maintaining its performance. Our results demonstrate that Hydra-PPO is a simple and promising solution for enabling more widespread usage of RLHF.

\end{abstract}

\blfootnote{Preprint.}


\section{Introduction}

Since ChatGPT, GPT-4, and Llama-2 family models entered the public sphere, they have impressed users with their ability to be helpful assistants for a surprising number of tasks \cite{bubeck2023sparks, qin2023chatgpt, zhao2023survey, laskar2023systematic,touvron2023llama}.  One key to their success, along with many other foundation models \cite{bommasani2022opportunities}, is model alignment through RLHF. Training a massive language model results in a network with a large amount of knowledge, however, it is not trained to \textit{discriminate within} that knowledge, which could cause undesired behaviour and possibly lead to societal harm \cite{10.1145/3442188.3445922}. Alignment aims to solve this issue by adjusting the model's behaviour and has become an integral part for creating safe and controllable foundation models \cite{bai2022constitutional, bai2022training}.  

While RLHF improves model alignment it is limited in usage, being both highly complex and demanding a massive amount of memory when loading and training multiple models during PPO  \cite{ouyang2022training, schulman2017proximal}.  Because the use of RLHF is in its infancy, there is a strong need to evaluate its variations in terms of speed and performance. 

To address this need, we delve into the training process and model architectures of standard RLHF-PPO. Through this investigation, we identify substantial opportunities for memory/computation cost reduction through the implementation of model-sharing between Reference/Reward Models and Actor/Critic Models.



\begin{table}[htbp]
  \centering
  \small
  \begin{tabular}{|c|c||c|c|c||c|c|c|}
    \hline
    \multirow{2}{*}{Method} & \multirow{2}{*}{\makecell{Batch \\ Size}} & \multicolumn{3}{c||}{  GPU Memory (GB) } &   \multicolumn{3}{c|}{  Latency per Sample (seconds) }   \\
    &  & Model & Activation & Total  & Inference  & Update &  Total\\ 
    \hline
    \hline

     PPO & 1   &  111.8*  &  101.3*  &  220* & - &   - & -\\
    LoRA-PPO & 1   &  53.2 &   12.5 &  68.0   & 17.23  &   1.52 &   18.75   \\
    J-Hydra-PPO  & 4  &  14.3  &   51.4 &   67.9   &  4.63  & 0.38 & \textbf{5.01} \\
    Hydra-PPO  & 4    &   15.9  &   52.8 &  71.1 &  4.88 &  1.59  &  \textbf{6.47} \\
    \hline
  \end{tabular}
  \captionsetup{skip=5pt}
  \caption{Comparison of Memory Usage and Run-Time between methods for Llama 7b on StackExchange per A100 80GB GPU. See Appendix \ref{apx:hparams} for details. \textbf{*}For Full Fine-Tuning PPO, memory usage is a scaled-up estimate.  }
  \label{tab:memory-comparison}
\end{table}

Given these findings, we propose Hydra-PPO to reduce the number of trained and static models in memory during PPO. We perform run-time and performance comparisons to show these memory savings can then be utilized to increase the training batch size, reducing the per-sample latency of PPO by up to 65\%.

\begin{figure}[h]
  \centering
  \includegraphics[scale=.7]{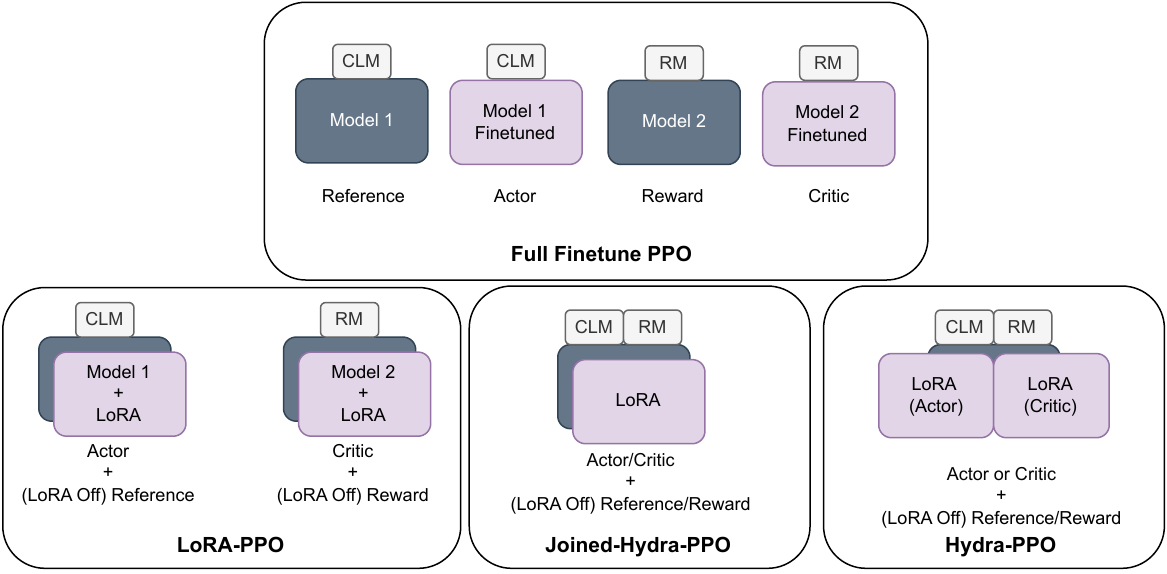}
  \caption{Models used in PPO methods. CLM indicates a Causal Language Modeling head, RM indicates a Reward Modeling head. Light purple weights are trained and dark blue weights are frozen.}
  \label{fig:comparison}
\end{figure}

\section{RLHF}
In this section, we first introduce the standard RLHF method \cite{ziegler2020finetuning, ouyang2022training, schulman2017proximal,rafailov2023direct}. 

\paragraph{Stage 1: Supervised Fine-Tuning (SFT)} an input LLM is trained using the standard causal language model training objective $\mathcal{L}_{\text{xent}}$ on a set of data $\mathcal{D}$, yielding language model $\pi_{\text{SFT}}$. We call this FFT-SFT when all parameters are trained, or LoRA-SFT when using LoRA \cite{hu2021lora}.

\paragraph{Stage 2: Reward Model (RM) Training} the head of a LLM is replaced with a scalar output. This model $r_{\phi}(x,y)$ is trained to predict human preference given a dataset of preference pairs with prompt $x$ and completion $y$. After training, the reward function is often normalized such that $\mathop{{}\mathbb{E}}_{x\sim\mathcal{D},y\sim\pi_{\text{SFT}}(y|x)}[r_\phi(x)] = 0$ to improve PPO stability. The reward model is trained with loss $ \mathcal{L}_R(r_\phi, \mathcal{D}) = -\mathop{{}\mathbb{E}}_{(x,y_w, y_l)\sim\mathcal{D}}[\text{log}(\,\sigma(r_\phi(x,\,y_w) - r_\phi(x,\,y_l)\,)] $, where $y_w$ is the "winning" answer as compared to $y_l$ for prompt $x$, according to the target alignment source.

\paragraph{Stage 3: PPO} $\pi_{\text{SFT}}$ and $r_{\phi}(x,y)$ are used to initialize and subsequently train an actor and critic with PPO \cite{schulman2017proximal, ouyang2022training}. During training, there are at minimum\footnote{Other models may be added \cite{ouyang2022training}; we stick to the most common and simplest setup in our paper.} four models used: 
\begin{itemize}
    \itemsep0em 
    \item \textbf{Reference:} $\pi_{\text{ref}}$, a frozen copy of $\pi_{\text{SFT}}$, used to prevent reward divergence.
    \item  \textbf{Actor:} called $ \pi_\theta$, the trained generative model or policy, initialized as a copy of $\pi_{\text{SFT}}$.
    \item \textbf{Reward:} a frozen copy of $r_{\phi}(x,y)$, used to calculate the reward of outputs from the Actor. 
    \item \textbf{Critic or Value Function:} $V(x,y)$, a copy of $r_{\phi}(x,y)$ trained to estimate sequence returns.
\end{itemize}

Using output probability ratio $r(\theta) = \frac{  \pi_\theta(y\: |\: x)   }{  \pi_{\text{old}}(y \: | \: x)   }$, PPO optimizes the surrogate objective $\mathcal{L}^\text{CLIP}(\theta) = \mathbb{E} [ \text{min}(r(\theta)\hat{A}, \text{clip}(r(\theta), 1-\epsilon, 1+\epsilon)\hat{A}  ]  $  . Generalized advantage estimation uses $V(x,y)$ to construct advantage estimates $\hat{A}$ from the reward \cite{schulman2018highdimensional, mnih2016asynchronous}. $V(x,y)$ is trained with squared-error loss on the returns. We use LoRA \cite{hu2021lora} on all linear layers of $ \pi_\theta$ and $V(x,y)$, which we call LoRA-PPO. We do not perform experiments with Full Fine-Tuning PPO due to its extreme cost.

\section{Hydra-RLHF}

We introduce Hydra-RLHF as a set of modifications to RLHF. We define a decoder-based model $\pi^{\text{hydra}}$ with two linear heads: 1) a head serves as the causal head, predicting the subsequent token for a sequence, and 2) another head serves as the reward model head, providing the immediate reward associated with the same input. Multi-headed models are well-explored both in general \cite{ruder2017overview, crawshaw2020multitask} and with respect to reinforcement learning \cite{mnih2016asynchronous, SilverHuangEtAl16nature, fletberliac2020merl}.

\paragraph{Stage 1: Hydra-SFT} \label{sec:crm} Using a similar dataset to standard RM training, $\pi^{\text{hydra}}$  is trained by optimizing $\mathcal{L}_{\pi^{\text{hydra}}}(x, y_w, y_l) = \mathcal{L}_{\text{xent}}(x, y_w) + \gamma\mathcal{L}_\theta(x, y_w, y_l)$, where $\gamma$ is a weighting multiplier. In practice, we find $\gamma = 0.1$ generally works well. We call this Hydra-FFT when training all parameters.

There are additional requirements for $\pi^{\text{hydra}}$ compared to regular RM or SFT fine-tuning. $\mathcal{L}_{\pi^{\text{hydra}}}(x, y_w, y_l)$ requires pairwise comparison data to train both heads, making standard SFT datasets unusable. Additionally, RM training can incorporate feedback from a list of rankings, e.g. $y_1 > y_2 > y_3$, by making pairs for all ranking combinations. For $\pi^{\text{hydra}}$, only pairs containing the sample with the best ranking should be considered to avoid training the SFT head on other samples. 


\paragraph{Dynamic LoRA} We introduce Dynamic LoRA as a simple and helpful technique to conserve memory usage in LoRA-PPO. Because $\pi_\theta$ and $\pi_{\text{ref}}$ are initialized as copies of $\pi_{\text{SFT}}$, training $\pi_\theta$ with LoRA \cite{hu2021lora} means the only difference between them is the LoRA weights. Rather than loading $\pi_{\text{SFT}}$ twice, $\pi_{\text{ref}}$ can be recovered from the actor by "turning off" LoRA. Thus, we define $\pi_{\text{ref}} \leftarrow \text{LO}(\pi_\theta)$, where $\text{LO}$ ignores any LoRA parameters. We add $ r_{\phi}(x,y) \leftarrow \text{LO}(V(x,y))$ for the Critic/Reward pair, saving about 20\% of memory while maintaining performance equivalent to LoRA-PPO.

\paragraph{Stage 2: Hydra-PPO} Two separate sets of LoRA weights are added to the same base model $\pi^{\text{hydra}}$, one set for the actor and one set for the critic, in order to create $\pi^{\text{RL-hydra}}_\theta$. When the actor is required, only the actor LoRA weights are used, and similarly for the critic. Utilizing dynamic LoRA, we define ($\pi^{\text{hydra}}_\text{ref}, r^{\text{hydra}}_{\phi}(x,y)) \leftarrow \text{LO}(\pi^{\text{RL-hydra}}_\theta)$. Only one full base model is required in memory during PPO, leading to similar overall memory usage to LoRA finetuning given the same batch size.

As an ablation study, we also include results of Joined Hydra-PPO or J-Hydra-PPO, which uses only one set of LoRA weights for both actor and critic. While this saves a small amount of memory and run-time, we find that it performs worse than Hydra-PPO. This is an interesting contrast to Hydra-SFT where joining the models does not affect performance.

\begin{table}[htbp]
  \centering
  \begin{tabular}{ccc}
    \toprule
    Method & \# of Static Models & \# of LoRA Weight Sets \\
    \midrule
    Full Fine-Tuning PPO & 4  &   0 (fully finetuned) \\
    LoRA-PPO & 4  &   2  \\
    Dynamic LoRA-PPO & 2  &   2  \\
    Joined Hydra-PPO & 1  &   1  \\
    Hydra-PPO & 1  &   2  \\
    \bottomrule
  \end{tabular}
  \captionsetup{skip=5pt}
  \caption{Summary of all PPO methods and number of models.}
  \label{tab:method-comparison}
\end{table}

\section{Experiments}

Results are presented across four datasets using Llama 7b \cite{touvron2023llama} or OPT 1.3b \cite{zhang2022opt}. 
We employ GPT-4 to evaluate model performance in general \cite{peng2023instruction, bai2022constitutional, kim2023aligning, liu2023geval}, and for the summarization task, we use also ROUGE scores\cite{lin-2004-rouge} .

In the empirical study, we evaluate five approaches: SFT, LoRA-PPO, Hydra-SFT, J-Hydra-PPO, and Hydra-PPO. Specifically, LoRA-PPO is initialized with the SFT model, while both J-Hydra-PPO and Hydra-PPO are initialized with the Hydra-SFT model.
All experiment hyperparameters are listed in Appendix \ref{apx:hparams}. Perplexity and RM accuracy before PPO is listed in Appendix \ref{apx:hparams}. Our code is forked from DeepSpeed-Chat \cite{yao2023deepspeedchat, 
 10.1145/3394486.3406703}.

The performance of PPO can be highly inconsistent due to its unstable nature and varying implementations \cite{engstrom2020implementation, rafailov2023direct, yuan2023rrhf, schulman2017proximal}. PPO can even reduce performance by exploiting flaws in the reward model, which we observe in the StackExchange dataset. 


\paragraph{Results Overview}



Tables \ref{tab:winrates-llama} and \ref{tab:winrates-opt} show the expected win-rates of each method against all other methods, as evaluated by GPT-4. The findings indicate that PPO outperforms SFT on average and Hydra-PPO similarly improves Hydra-SFT. The specific win-rates per dataset are provided in detail. The performance of SFT and Hydra-SFT are comparable, suggesting that combining the RM and SFT objectives within a single model does not consistently lead to improvements or hinder the generation performance across different tasks. 

Both Hydra-PPO and LoRA-PPO improve over their respective base models, however, Hydra-PPO achieves better alignment than LoRA-PPO for Llama 7b. This may be explained by the better Reward model from Hydra-SFT which enables overall better PPO performance. The detailed accuracy of the RM models in SFT and Hydra-SFT is shown in Appendix \ref{apx:pre-ppo}. Overall, the study indicates that PPO improves model alignment and there is potential for further enhancing PPO performance by improving the RM.  

For Learning to Summarize, we additionally evaluate their performance using ROUGE scores in Table \ref{tab:rouge-l2s}, and these results consistently align with the findings presented in Table \ref{tab:winrates-llama}. An interesting observation is that the SFT-based approach typically yields better precision performance, whereas PPO-based methods substantially enhance recall. This trend could potentially be attributed to the encouragement of longer text generation during the PPO stage.


\paragraph{Joined-Hydra-PPO Underperformance} J-Hydra-PPO, which uses only one set of LoRA weights for actor and critic, performs significantly worse than two separate sets (Hydra-PPO). We speculate this is due to combining actor and critic model amplified the unstable nature of PPO \cite{engstrom2020implementation, rafailov2023direct, yuan2023rrhf, schulman2017proximal}. Since J-Hydra-PPO is more memory and computation efficient than Hydra-PPO, we hope future work may improve its performance.


\begin{table}[htbp]
  \centering
  \begin{tabular}{|c||c|c|c|c|c|}
    \hline
    Method & GPT-4-LLM & \makecell{Open-Source\\  Assistant} & \makecell{Learning to \\ Summarize} & StackExchange  & \textbf{Average} \\ 
    \hline
    \hline
    
    SFT & 48.18     &48.35     & 45.95   & 51.73  &  48.55 \\
    LoRA-PPO &  48.8  & 49.03     & 55.48    & 49.4  &  \textbf{50.68}  \\
    \hline
    Hydra-SFT & 48.48  &  49.65           & 42.63      & 53.23  &  48.50     \\
    J-Hydra-PPO  &  50.43     & \textbf{52.05}     & 43.13   & 40.38 & 46.50   \\
    Hydra-PPO  & \textbf{54.13}  & 51          & \textbf{61.58}   & \textbf{55.38} &  \textbf{55.52} \\
    \hline
  \end{tabular}
  \captionsetup{skip=5pt}
  \caption{Llama 7b expected aggregate win-rate per method. We measure total wins and ties for each method against all other methods, then use this to calculate expected win-rate. }
  \label{tab:winrates-llama}
\end{table}

\begin{table}[htbp]
  \centering
  \begin{tabular}{|l|lll|lll|}
    \hline
    \multirow{2}{1em}{Model} & \multicolumn{3}{|c|}{ROUGE-1}  & \multicolumn{3}{|c|}{ROUGE-L}  \\
     & Precision & Recall & F-Measure &  Precision & Recall & F-Measure \\
    \hline
    \hline
    SFT &  90.69    &  13.12  & 21.69  &  75.56  & 11.35  & 18.59  \\
    LoRA-PPO &  88.93  &  14.70   & 23.95  & 71.46  & 12.25  &   19.77  \\
    \hline
    Hydra-SFT &  87.86  & 13.27  & 21.42 & 72.92 & 11.42 &  18.27 \\
    
    J-Hydra-PPO &  84.13  & 16.93  & 25.00 & 70.82 & 14.92 &  21.81  \\
    Hydra-PPO &  88.91  & 19.21  & \textbf{29.31}  & 72.45 & 16.43 & \textbf{24.73} \\
    \hline
  \end{tabular}
  \captionsetup{skip=5pt}
  \caption{Llama 7b ROUGE-1 and ROUGE-L scores for all models on the Learning to Summarize dataset.}
  \label{tab:rouge-l2s}
\end{table}

\begin{figure}[htbp]
  \centering
  \includegraphics[scale=.6]{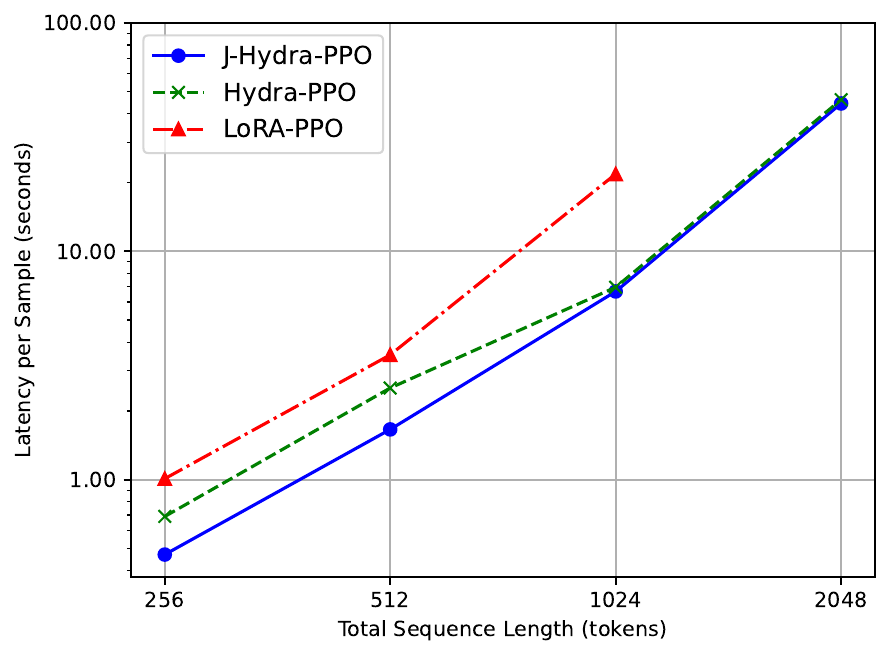}
  \caption{Latency (seconds) per Sample per PPO method as sequence length increases. Both axes use log scaling. LoRA-PPO is unable to fit in memory for our setup for context length 2048. See Appendix \ref{apx:hparams} for details.} 
  \label{fig:throughput}
\end{figure}

\paragraph{Throughput Comparison} Figure \ref{fig:throughput} shows there is a roughly linear relationship between throughput and sequence length in log-space for all methods. Latency is measured as a sum of inference latency and parameter update latency per sample during PPO. As we can see from the figure, Hydra-PPO saves exponentially more time as sequence length increases. We increase batch size to max out memory usage for all methods, but use gradient accumulation to ensure the effective total batch size is the same. Hydra-PPO and J-Hydra-PPO converge at sequence length 1024 as the inference increase overtakes update latency. Table \ref{tab:memory-comparison} shows a detailed comparison for a specific experiment. 

\subsection{GPT-4-LLM}

GPT-4-LLM \cite{peng2023instruction} consists of instruction-following prompts with responses sampled from multiple foundation models, including GPT-4, GPT-3.5, and OPT-IML. The responses of each model are ranked by GPT-4. We pair only the highest-scoring response with each other response. To our knowledge, we are the first to attempt full RLHF (including PPO) on this dataset. Overall, we observe the most consistent and well-behaved training runs with GPT-4-LLM.

\begin{table}[H]
  \centering
\begin{tabular}{|l||l|l|l|l|}
\hline
-              &  Hydra-FFT     &  LoRA-PPO                & J-Hydra-PPO          & Hydra-PPO      \\ 
\hline
\hline
SFT               &  40.6 / \textbf{43.8}   &  \textbf{43.6} / 43.4    &  43.2 / \textbf{46.8}    &  39.0 / \textbf{47.0}            \\
Hydra-FFT         & -                  &  43.4 / \textbf{45.2}   & 40.8 / \textbf{44.0}    &   38.8 / \textbf{49.2}         \\
LoRA-PPO           & -                  & -                  &  40.8 / \textbf{44.4}  &  40.0 / \textbf{47.6}    \\
J-Hydra-PPO         & -                  & -                  & -                  &  38.6 / \textbf{45.6}       \\
\hline              
\end{tabular}
  \captionsetup{skip=5pt}
  \caption{Llama 7b GPT-4-LLM win-rates as judged by GPT-4. Results in each cell are presented as "Row Win \% / Column Win \%" with the remainder being ties.}
  \label{tab:all-gpt4llm}
\end{table}

\subsection{Open-Source Assistant Datasets}

We perform RLHF on the default data for DeepSpeed-Chat \cite{yao2023deepspeedchat,10.1145/3394486.3406703}. At the time of writing, these datasets include "Dahoas/rm-static", "Dahoas/full-hh-rlhf", "Dahoas/synthetic-instruct-gptj-pairwise" and "yitingxie/rlhf-reward-datasets", all hosted on HuggingFace. We call the combination "Open-Source Assistant Datasets". These are various open-source ChatBot or Assistant style datasets, with one including Helpful \& Harmless \cite{bai2022training}. We train on them without modification.

\begin{table}[htbp]
  \centering
\begin{tabular}{|l||l|l|l|l|}
\hline
-              &  Hydra-FFT     & LoRA-PPO                & J-Hydra-PPO          & Hydra-PPO      \\ 
\hline
\hline
SFT               &   42.2/ \textbf{44.4}   &      41.7 / \textbf{42.2}     &     40.6 / \textbf{45.2}         &    39.4 / \textbf{45.4}            \\
Hydra-FFT         & -                  &   \textbf{45.4} / 38.4        &  37.8 / \textbf{45.0}   &   38.6 / \textbf{43.6}             \\
LoRA-PPO               & -                  & -        &  43.0 / \textbf{44.4}   &  42.6 / \textbf{42.8}     \\
J-Hydra-PPO         & -                  & -                  & - &  \textbf{45.6}  / 42.4      \\
\hline              
\end{tabular}
  \captionsetup{skip=5pt}
  \caption{Llama 7b Open-Source Assistant Datasets win-rates as judged by GPT-4. Results in each cell are presented as "Row Win \% / Column Win \%" with the remainder being ties.}
  \label{tab:all-oad}
\end{table}

\subsection{ Learning to Summarize }

The Reddit TL;DR dataset \cite{volske-etal-2017-tl} has been previously used in multiple RLHF works \cite{stiennon2022learning, ziegler2020finetuning}. We use the dataset as modified by \cite{stiennon2022learning}, where each prompt only contains one preference completion pair. 


\begin{table}[H]
  \centering
\begin{tabular}{|l||l|l|l|l|}
\hline
-              &  Hydra-FFT     & LoRA-PPO                & J-Hydra-PPO          & Hydra-PPO      \\ 
\hline
\hline
SFT               &   \textbf{41.0} / 38.4       &   31.6 / \textbf{44.8}   &    \textbf{41.0} / 32.8     &    31.4 / \textbf{47.2}             \\
Hydra-FFT         & -    &      33.4 / \textbf{46.4}       &  \textbf{37.8} /  36.4   &   33.8 / \textbf{42.4}         \\
LoRA-PPO               & -                  & -                  & \textbf{51.8} / 30.0  &  \textbf{42.6} / 36.8    \\
J-Hydra-PPO         & -                  & -                  &   -   &   23.7 / \textbf{52.6}    \\
\hline              
\end{tabular}
  \captionsetup{skip=5pt}
  \caption{Llama 7b Learning to Summarize win-rates as judged by GPT-4. Results in each cell are presented as "Row Win \% / Column Win \%" with the remainder being ties.}
  \label{tab:gpt4-learn2sum}
\end{table}

\subsection{StackExchange}

The StackExchange \cite{h4stackexchange} dataset has previously been used to train StackLlama via RLHF \cite{beeching2023stackllama}. Each data sample consists of one question with multiple completions ranked by votes from users. We re-create this experiment with 150k samples from StackExchange, with a change in that we pair only the best answer with up to 3 other answers. This is done to avoid over-training on the best sample in Hydra-SFT, but in addition, we find that the most up-voted answers are on average longer than the other answers, leading to trivial reward models.

StackExchange is the most difficult dataset we test, containing extremely diverse and specific questions. During PPO, models often learn to repeat their answers. Despite multiple attempts, both the PPO and J-Hydra-PPO models encounter this issue while Hydra-PPO does not.

\begin{table}[htbp]
  \centering
\begin{tabular}{|l||l|l|l|l|}
\hline
-              &  Hydra-FFT     & LoRA-PPO                & J-Hydra-PPO          & Hydra-PPO      \\ 
\hline
\hline
SFT               &     41.2 / \textbf{42.4}            &    \textbf{46.4} / 42.0    &        \textbf{51.8} / 35.0       &    42.4 / \textbf{48.6}            \\
Hydra-FFT         & -                  &    \textbf{46.4} / 43.2       &    \textbf{54.2} / 32.4    &     45.2 / \textbf{45.6}              \\
LoRA-PPO               & -                  & -                  &  \textbf{52.6} / 34.8   &  36.8 / \textbf{51.8}    \\
J-Hydra-PPO         & -                  & -                  & - &  35.2 / \textbf{56.6}    \\
\hline              
\end{tabular}
  \captionsetup{skip=5pt}
  \caption{Llama 7b StackExchange win-rates as judged by GPT-4. Results in each cell are presented as "Row Win \% / Column Win \%" with the remainder being ties.}
  \label{tab:all-StackExchange}
\end{table}

\subsection{Changing Model Size and Family}

We extend our experimentation to explore the SFT and PPO approaches using the OPT-1.3b model. For this model, we find that Hydra-SFT performs worse than the SFT model. Additionally, we find LoRA-PPO has better overall alignment than Hydra-PPO for OPT-1.3b. We speculate this difference to be due to the capacity of the model. For the smaller 1.3b model, combining language and reward models may be more difficult. Overall, we observe the same trend in increased performance after PPO and Hydra-PPO over their respective base models.

\begin{table}[H]
  \centering
  \begin{tabular}{|c||c|c|c|}
    \hline
    Method & GPT-4-LLM & \makecell{Open-Source\\  Assistant} & \textbf{Average} \\ 
    \hline
    \hline
    
    SFT &  45.65   &   52.5   &   49.08   \\
    LoRA-PPO &  \textbf{59.5}  &    53.7   &     \textbf{56.6}        \\
    \hline
    Hydra-SFT & 44.4 &   42.58  &    43.49         \\
    J-Hydra-PPO  &   48.2    &    46.78  &   47.49   \\
    Hydra-PPO  &  50.2  &    \textbf{54.45}   &  \textbf{52.33}  \\
    \hline
  \end{tabular}
  \captionsetup{skip=5pt}
  \caption{OPT 1.3b expected aggregate win-rate per method. We measure total wins and ties for each method against all other methods, then use this to calculate expected win-rate.}
  \label{tab:winrates-opt}
\end{table}

\begin{table}[htbp]
  \centering
\begin{tabular}{|l||l|l|l|l|}
\hline
-              &  Hydra-FFT     &  LoRA-PPO                & J-Hydra-PPO          & Hydra-PPO      \\ 
\hline
\hline
SFT               &  \textbf{52.2} / 36.6   &  41.2 / \textbf{44.0} &  \textbf{52.6} / 39.4         &     43.2 / \textbf{49.2}            \\
Hydra-FFT         & -                  &  35.8 / \textbf{53.4}     &  42.4 / \textbf{49.8}  &   37.8 / \textbf{56.6}            \\
LoRA-PPO           & -                  & -                  &   \textbf{50.8} / 41.2    &  46.2 / \textbf{46.6} \\
J-Hydra-PPO         & -                  & -                  & -                  &   39.2 / \textbf{49.6}    \\
\hline              
\end{tabular}
  \captionsetup{skip=5pt}
  \caption{OPT 1.3b GPT-4-LLM win-rates as judged by GPT-4. Results in each cell are presented as "Row Win \% / Column Win \%" with the remainder being ties.}
  \label{tab:all-gpt4llm-1.3b}
\end{table}

\begin{table}[htbp]
  \centering
\begin{tabular}{|l||l|l|l|l|}
\hline
-              &  Hydra-FFT     & LoRA-PPO                & J-Hydra-PPO          & Hydra-PPO      \\ 
\hline
\hline
SFT               &    \textbf{45.0} / 44.0  &   31.8 / \textbf{54.8}      &     41.2 / \textbf{47.4}   &   39.8 / \textbf{50.0}          \\
Hydra-FFT         & -                  &    34.6 / \textbf{53.0}        &   32.0 / \textbf{54.6}  &   42.4 / \textbf{44.2}            \\
LoRA-PPO               & -                  & -                  & \textbf{53.0} / 35.8    &  \textbf{52.0} / 33.6    \\
J-Hydra-PPO         & -                  & -                  & - &   39.0 / \textbf{47.0}    \\
\hline              
\end{tabular}
  \captionsetup{skip=5pt}
  \caption{OPT 1.3b Open-Source Assistant Datasets Preference as judged by GPT-4. Results in each cell are presented as "Row Win \% / Column Win \%" with the remainder being ties.}
  \label{tab:all-oad-1.3b}
\end{table}

\section{Related Works}

\paragraph{Aligning to Human Preference} Foundation models have begun to emerge as all-purpose language models \cite{bommasani2022opportunities} which may be used without any domain adaptation \cite{wei2022emergent, bubeck2023sparks, radford2019language}. While these models clearly contain a large amount of knowledge and ability, they may contain unintended bias or respond in unintended ways to input questions from a user. Model alignment is the problem of slightly modifying these models to interact with humans in a specific manner.

Human preference is difficult to quantify (and often inconsistent \cite{rafailov2023direct, ouyang2022training}), making model alignment an open research area \cite{leike2018scalable}. By assuming that classification is easier than generation, it is possible to train a reward model on a dataset of human preference labels. Such a reward model may then be used to guide other models towards aligning to human preference, improving performance in a nontrivial way over Supervised fine-tuning (SFT) throughout many domains \cite{christiano2023deep, ziegler2020finetuning, bai2022training, stiennon2022learning, bai2022constitutional, glaese2022improving, wu2021recursively}. Recently, this concept has exploded in popularity due to the success of InstructGPT and subsequent improvements in ChatGPT and GPT-4 \cite{ouyang2022training} which have delivered undeniably strong and human-like interactions in a variety of domains. 

Other forms of feedback have been attempted due to the high cost of hiring humans to grade inputs. Now that massive foundation models exist, multiple works have attempted to use their feedback to train or evaluate other models \cite{peng2023instruction, bai2022constitutional, kim2023aligning, liu2023geval,gunasekar2023textbooks,eldan2023tinystories}. 



\paragraph{Alignment during Supervised Fine-Tuning (SFT)} Due to the complexity and high cost of PPO, some recent works have sought to replace the training process of PPO while retaining its benefits. Notably, RAFT \cite{dong2023raft}, RRHF \cite{yuan2023rrhf}, PRO \cite{song2023preference}, and DPO \cite{rafailov2023direct} are recent methods which combine preference data in some way with supervised fine-tuning. The former two are inspired by best-of-\textit{n} sampling methods \cite{cobbe2021training, askell2021general, nakano2022webgpt}, while the latter two seek to wholly replace PPO by re-framing the supervised training objective.

Hydra-SFT shares similarities with these approaches by integrating ranked feedback into supervised fine-tuning. However, our work is orthogonal to these methods, aiming not to replace RLHF, but rather to make it more widely usable. 

\paragraph{Dataset Formation} Hydra-RLHF requires that the SFT and RM training datasets be the same. Previous works have found issues in over-fitting one of the heads when data is imbalanced \cite{ziegler2020finetuning, stiennon2022learning}. Our experiments use datasets with pairwise comparisons for each sample so we find this over-fitting is not an issue, however, Hydra-RLHF could be extended to handle exceptions when data is limited.

\paragraph{Reward Model Size} In RLHF, the reward model can be smaller than the language model. We keep these models the same size to make performance comparisons fair. In applied usage, Hydra-RLHF comparatively saves less memory when standard RLHF uses a smaller reward model, however, this is also an advantage for Hydra-RLHF; it uses a larger reward model for less training cost.

\section{Conclusion}

We have performed a comparative study which analyzes the performance of different approaches to model alignment as graded by GPT-4. We find that LoRA-PPO improves alignment over FFT but is costly to run. We introduce Hydra-RLHF as a method to save memory during PPO while maintaining performance, which consists of two major parts: combining reference and reward models, and dynamically switching the active LoRA module during PPO. With the excess memory, Hydra-RLHF may use a higher batch size and therefore train with up to 65\% faster per-sample latency. Hydra-RLHF opens up the possibility for the community to apply RLHF for a wider variety of models and applications. We also see potential for future improvements, notably, balancing the SFT and RM datasets, improving performance of J-Hydra-PPO, and improving PEFT methods for RLHF.

\section*{Acknowledgments}
Thank you to Vladimir Fomenko and Jialei Chen for helpful discussions. 

\bibliographystyle{unsrt}  
\bibliography{references}  

\newpage

\appendix

\section{Algorithm Pseudocode}

We present pseudocode for J-Hydra-RLHF and Hydra-RLHF to aid in their implementation.

\begin{algorithm}
\caption{J-Hydra-PPO}\label{alg:j-hydra-ppo}
\begin{algorithmic}
\For{iteration=1,2,...}

\For{actor=1,2,...N}
\State Generate sequence using policy $\pi^{\text{hydra}}_{\theta_\text{old}}$ for $T$ sequence length steps, retaining critic head values for the last step
\State Turn LoRA off for $\pi^{\text{hydra}}_{\theta_\text{old}}$ and use it to compute reward and reference log probabilites
\State Use reward and reference log probability to calculate $\hat{A}$
\EndFor

\State Optimize surrogate $\mathcal{L}$ wrt $\pi^{\text{hydra}}_\theta$, with $K$ epochs and minibatch size $M \leq NT$
\State $\pi^{\text{hydra}}_{\theta_\text{old}} \leftarrow \pi^{\text{hydra}}_\theta$
\EndFor

\end{algorithmic}
\end{algorithm}

\begin{algorithm}
\caption{Hydra-PPO}\label{alg:hydra-ppo}
\begin{algorithmic}
\For{iteration=1,2,...}

\For{actor=1,2,...N}
\State Generate sequence using policy $\pi^{\text{hydra}}_{\theta_\text{old}}$ for $T$ sequence length steps
\State Turn on the critic LoRA for $\pi^{\text{hydra}}_{\theta_\text{old}}$ and use it to calculate critic values on the generated sequence
\State Turn LoRA off for $\pi^{\text{hydra}}_{\theta_\text{old}}$ and use it to compute reward and reference log probabilites
\State Use reward and reference log probability to calculate $\hat{A}$
\EndFor

\State Optimize surrogate $\mathcal{L}$ wrt $\pi^{\text{hydra}}_\theta$, with $K$ epochs and minibatch size $M \leq NT$
\State $\pi^{\text{hydra}}_{\theta_\text{old}} \leftarrow \pi^{\text{hydra}}_\theta$
\EndFor

\end{algorithmic}
\end{algorithm}

\section{Experiment Details} \label{apx:hparams}

\paragraph{Setup} All hyper-parameters used are listed in Tables \ref{tab:hparams} and \ref{tab:hparams-1.3b}. All experiments ran on a 8xA100 80GB node. and batch size is measured per-GPU, making total batch size = 8 * batch size * gradient accumulation steps. All experiments use DeepSpeed ZeRO-1 \cite{10.1145/3394486.3406703, DeepSpeed-Chat}. All experiments use a cosine decaying learning rate scheduler. 

We fix a common set of parameters. For all experiments, for SFT, RM, and Hydra-SFT, weight decay is set to 0.1. Reward models are normalized between (-1, 1). When using LoRA, SFT models use a LoRA rank of 128. Across all SFT, RM, and Hydra-SFT experiments, we began by sweeping learning rates \{5e-5, 5e-6, and 5e-7\} for 4 epochs, then adjusted parameters accordingly. We use the model with the best validation performance across epochs. 

For all experiments, PPO, J-Hydra-PPO, and Hydra-PPO share the following: KL-Divergence $\beta$ = 0.02, GAE $\gamma$ = 1.0, GAE $\lambda$ = 0.95, warmup steps = 100, LoRA rank for actor and critic = 128, and all are run for 1 epoch. We track the mean of the reward from the last 20 steps, and use the model with the best reward throughout training. 

\paragraph{Run-Time Measurement} All run-time numbers were obtained from training Llama 7b on StackExchange. For Table \ref{tab:memory-comparison}, we use a total sequence length of 800 tokens. Total GPU memory exceeds the sum of Model and Activation due to other stored tensors used in training, while Total Latency per Sample is a direct sum of the two stages of PPO. Memory usage was measured with PyTorch. Because Full PPO overflows our setup, we repeat the experiment using OPT 1.3b, and use the ratio between LoRA-PPO on Llama 7b to scale the numbers up. 

\paragraph{PPO Hyperparameter Stability} For all PPO runs, we ensure reward increases during training and take the point of the highest reward without extreme and obvious divergence. In general, we find J-Hydra-PPO to be highly unstable, taking multiple attempts to find solid hyperparameters. Hydra-PPO is generally as stable as LoRA-PPO.

\begin{table}[H]
  \centering
  \small
  \begin{tabular}{ccccc}
    \toprule
    Hyperparameter & GPT-4-LLM & Learning to Sum. & Open-Source Assistant & StackExchange  \\
    \midrule
    Prompt Sequence Length &  256   & 512  &   256  &  400  \\
    Completion Sequence Length &   256  &  256 &  256   &  400  \\
    \midrule
    \multicolumn{5}{c}{SFT} \\
    \midrule
    Learning Rate &  1e-5   & 1e-4  &  1e-3    &  1e-4  \\
    Batch Size  & 4    & 3    &  4   &  1  \\
    Gradient Accum. Steps  &  1   & 3  &  1   &  6  \\
    Epochs &   3  & 7  &  7   &   4 \\
    \midrule
    \multicolumn{5}{c}{RM} \\
    \midrule
    Learning Rate &  5e-5   & 5e-5  &  5e-4   &  1e-6  \\
    Batch Size  &   4  & 1  &   3  &   1 \\
    Gradient Accum. Steps  & 1    & 4  & 1    &  12  \\
    Epochs &   3  &  3 & 6    &  3  \\
    \midrule
    \multicolumn{5}{c}{Hydra-SFT} \\
    \midrule
    Learning Rate &  3e-6  & 5e-7  &   1e-5   &  5e-7  \\
    Reward Head Multiplier & 0.1  & 0.1  &    0.1   &  0.07  \\
    Batch Size  &  4 &  1  &   4    &  1  \\
    Gradient Accum. Steps  & 1   & 10  &  1    &  6  \\
    Epochs &   3  & 4  &   4  &  6  \\
    \midrule
    \multicolumn{5}{c}{LoRA-PPO} \\
    \midrule
    Actor Learning Rate    &  5e-4   &  8e-4  &  5e-4   & 5e-4   \\
    Critic Learning Rate   &  5e-5   & 1e-4  &  5e-5   &  5e-5  \\
    Generation Batch Size  &  6   & 2  &  6   &  1  \\
    Mini-Batch Size  &   8   &  4 &  8   &  3  \\
    Gradient Accum. Steps  &  5   & 30  &  3   & 25   \\
    \midrule
    \multicolumn{5}{c}{J-Hydra-PPO} \\
    \midrule
    Actor Learning Rate &   5e-3  &  2.5e-4  &  1e-4   &  6e-4  \\
    Critic Loss Multiplier &  0.1   & 3  &  0.1   &  3  \\
    Generation Batch Size  &   6  &  2  & 6    &  1  \\
    Mini-Batch Size  &  8   &  4  &  8   &  3  \\
    Gradient Accum. Steps  &  5   & 50  &  3   &  25  \\
    \midrule
    \multicolumn{5}{c}{Hydra-PPO} \\
    \midrule
    Actor Learning Rate &  5e-3   &  5e-4 &  5e-5   &  7e-4  \\
    Critic Learning Rate &  5e-5   & 5e-5  &  5e-6   &  8e-4  \\
    Generation Batch Size  &   6  & 2  &  6   &  1  \\
    Mini-Batch Size  &  8   & 4  &  8   &   3 \\
    Gradient Accum. Steps  &  5   & 30  &   3  &  25  \\
    \bottomrule
  \end{tabular}
  \captionsetup{skip=5pt}
  \caption{Hyperparameters for Llama 7b on all datasets.}
  \label{tab:hparams}
\end{table}

\begin{table}[H]
  \centering
  \small
  \begin{tabular}{ccc}
    \toprule
    Hyperparameter & GPT-4-LLM &  Open-Source Assistant  \\
    \midrule
    Prompt Sequence Length &  256    &   256  \\
    Completion Sequence Length &   256 &  256    \\
    \midrule
    \multicolumn{3}{c}{SFT}  \\
    \midrule
    Learning Rate &  5e-5    &  1e-6 \\
    Batch Size &  4   & 4 \\
    Gradient Accum. Steps  &  1   & 1 \\
    Epochs &  3   & 4 \\
    \midrule
    \multicolumn{3}{c}{RM} \\
    \midrule
    Learning Rate  &   5e-5     &  1e-5 \\
    Batch Size  &  4   & 4 \\
    Gradient Accum. Steps  &  4   &  4\\
    Epochs  &   3  & 3 \\
    \midrule
    \multicolumn{3}{c}{Hydra-SFT} \\
    \midrule
    Learning Rate & 3e-6 &  5e-5 \\
    Reward Head Multiplier &  .1   & .1   \\
    Batch Size  & 4    & 4  \\
    Gradient Accum. Steps & 1    &1  \\
    Epochs  &  3   &  4  \\
    \midrule
    \multicolumn{3}{c}{LoRA-PPO} \\
    \midrule
    Actor Learning Rate    &  2.5e-4   &  5e-4    \\
    Critic Learning Rate   &  5e-5    &  5e-5     \\
    Generation Batch Size  &  6    &  6   \\
    Mini-Batch Size  &   8   &  8   \\
    Gradient Accum. Steps  &  5     &  3      \\
    \midrule
    \multicolumn{3}{c}{J-Hydra-PPO} \\
    \midrule
    Actor Learning Rate &   5e-4    &  5e-4     \\
    Critic Loss Multiplier &  0.1    &  0.1   \\
    Generation Batch Size  &   6   & 6      \\
    Mini-Batch Size  &  8      &  8      \\
    Gradient Accum. Steps  &  5     &  3     \\
    \midrule
    \multicolumn{3}{c}{Hydra-PPO} \\
    \midrule
    Actor Learning Rate &  5e-4    &  5e-4     \\
    Critic Learning Rate &  5e-5     &  1e-5     \\
    Generation Batch Size  &   6     &  6     \\
    Mini-Batch Size  &  8     &  8    \\
    Gradient Accum. Steps  &  5    &   3    \\
    \bottomrule
  \end{tabular}
  \captionsetup{skip=5pt}
  \caption{Hyperparameters for OPT 1.3b on all datasets.}
  \label{tab:hparams-1.3b}
\end{table}

\section{GPT-4 Grading}

Significant consideration was given to the prompt for grading model responses. We experimented with one prompt which contained two answers to decide from (e.g. "out of answer one or two, which is better"). However, we found that simply switching the order of the answers would switch the preferred answer by a large margin. 

Instead, we find the most consistent evaluation method to be grading each output on a 0-9 scale with step-by-step reasoning, and then comparing these grades side-by-side. Each output is graded 3 times at temperature 1 and the results are averaged for a final score.

When grading, we perform cleanup on answers from all models: removing EOT tokens, stopping generation upon detection of the model speaking for the "other side" (i.e. begins writing a new instruction and response for Open-Datasets questions), and stopping generation upon detection of a 5-gram phrase sequence already found in the sequence. We evaluate using 500 samples from the validation set of each dataset.

To grade outputs on the GPT-4-LLM, Open-Datasets, and StackExchange datasets, we use the following prompt:

\noindent\fbox{%
    \parbox{\textwidth}{%
System Prompt: 
\newline
You are an unbiased and harsh critic who judges the quality of answers from an assistant.
\newline
\newline
User Message:
\newline
Given a question from a user, rate the given answer from an assistant with an integer score out of 9. Your rating can be any number from 0, 1, 2, 3, 4, 5, 6, 7, 8, or 9. First, give an extremely short, 1-sentence evaluation of the answer. Do not give bullet points. Do not write more than one sentence. Only write one total sentence evaluating the answer. A better answer is one that is more helpful, answers the question more precisely, is more concise, is grammatically correct, is kinder and more friendly, and does not contain any rude or confrontational speech. Give your answer in exactly this format and then end the answer:
\newline
\newline
Let us think step-by-step.
\newline
[1 sentence evaluating the quality of the answer]
\newline
The integer score out of 9 for this answer is: X
\newline
\newline
Here is the original conversation and then question from the user:
\newline
\newline
\textit{Insert conversation and question}
\newline
\newline
Here is the answer from the assistant:
\newline
\newline
\textit{Insert answer to judge}
    }%
}

To grade outputs from Learning to Summarize, we use the following prompt:

\noindent\fbox{%
    \parbox{\textwidth}{%
System Prompt: 
\newline
You are an unbiased and harsh critic who judges the quality of summaries of a piece of text.
\newline
\newline
User Message:
\newline
Given a summary of a piece of text, rate the quality of the summary with an integer score out of 9. Your rating can be any number from 0, 1, 2, 3, 4, 5, 6, 7, 8, or 9. First, give an extremely short, 1-sentence evaluation of the summary. Do not give bullet points. Do not write more than one sentence. Only write one total sentence evaluating the summary. A better summary is one that is more concise, captures more of the sentiment of the input text, does not contain any false information about the input, and copies minimal text directly from the input. Give your answer in exactly this format and then end the answer:
\newline
\newline
Let us think step-by-step.
\newline
[1 sentence evaluating the quality of the summary]
\newline
The integer score out of 9 for this answer is: X
\newline
\newline
Here is the original text:
\newline
\newline
\textit{Insert original text}
\newline
\newline
Here is the summary of the text:
\newline
\newline
\textit{Insert summary to judge}
    }%
}

\newpage

\section{Reward Model Grading} \label{apx:rm-eval}

We use GPT-4 as our primary evaluator, however, it is possible to use a reward model as the evaluator. Table \ref{tab:rm-eval} shows the win-rates of FFT-SFT vs. LoRA-PPO using the separated RM, and Hydra-SFT vs. J-Hydra-PPO and Hydra-SFT vs. Hydra-PPO using the Hydra-RM. This table shows the success of each PPO method in optimizing their actors towards their reward model. Overwhelmingly, the RMs rate each PPO model as strongly superior to its respective input model.

\begin{table}[htbp]
  \centering
  \begin{tabular}{cccc}
    \toprule
    Dataset & \makecell{LoRA-PPO} &  J-Hydra-PPO   & Hydra-PPO  \\
    \midrule
    \multicolumn{4}{c}{OPT 1.3b} \\
    \midrule
    GPT-4-LLM & 19.4 / \textbf{70.2}  & 20.4 / \textbf{79.6} & 13.4 / \textbf{86.6} \\
    Open-Source Assistant & 20.4 / \textbf{78.6}  & 25.8 / \textbf{41.2}  & 26.8 / \textbf{40.8} \\
    \midrule
    \multicolumn{4}{c}{Llama 7b} \\
    \midrule
    GPT-4-LLM & 13.6 / \textbf{82.6} & 36.4 / \textbf{62.2} & 27.8 / \textbf{69.2} \\
    Open-Source Assistant  & 12.4 / \textbf{14.8}   & 15.0 / \textbf{40.2} & 21.6 / \textbf{64.4}\\
    Learning to Summarize & 34.8 / \textbf{53.8}  & \textbf{44.4} / 31.0 & \textbf{54.6} / 40.8\\
    StackExchange  & 10.2 / \textbf{89.8}  & \textbf{54.0} / 45.6  & 11.0 / \textbf{89.0}\\
    \bottomrule
  \end{tabular}
  \captionsetup{skip=5pt}
  \caption{Win Rates of each PPO method versus its input model (FFT-SFT / LoRA-PPO, Hydra-SFT / J-Hydra-PPO, and Hydra-SFT / Hydra-PPO). Results in each cell are presented as "Base Wins / PPO wins" with the remainder being ties.}
  \label{tab:rm-eval}
\end{table}

\section{LoRA-SFT vs.  LoRA-PPO} \label{apx:lora-sft-vs-ppo}

For our experiments, we use FFT-SFT for both standard and hydra models. Table \ref{tab:lora-vs-ppo} shows results of using LoRA-SFT against LoRA-PPO, as graded by GPT-4. Here,  LoRA-PPO uses FFT as the input model. We find LoRA SFT under-performs compared to FFT. While efficient, LoRA does not always match FFT-SFT performance \cite{hu2021lora, sun2023comparative, pu2023empirical, lialin2023stack}. More investigation into PEFT \cite{peft} for alignment is required. If a method like \cite{lialin2023stack} could improve LoRA-SFT for alignment, the entire RLHF process could be done with roughly the same footprint as LoRA-SFT.

\begin{table}[htbp]
  \centering
  \begin{tabular}{cc}
    \toprule
    Dataset & LoRA-SFT / LoRA-PPO \\
    \midrule
    \multicolumn{2}{c}{OPT 1.3b} \\
    \midrule
    GPT-4-LLM & 42.0 / \textbf{45.2} \\
    Open-Source Assistant  & 29.2 / \textbf{61.0}  \\
    \midrule
    \multicolumn{2}{c}{Llama 7b} \\
    \midrule
    GPT-4-LLM & 35.2 / \textbf{53.0}    \\
    Open-Source Assistant   &  \textbf{47.7} / 42.1   \\
    Learning to Summarize  & 28.2 / \textbf{50.7 }  \\
    StackExchange  & \textbf{45.9} / 43.1  \\
    \bottomrule
  \end{tabular}
  \captionsetup{skip=5pt}
  \caption{Win Rates of LoRA-SFT vs. LoRA-PPO (with FFT-SFT base) on all datasets using GPT-4 as a judge. Results in each cell are presented as "LoRA-SFT Wins / LoRA-PPO wins" with the remainder being ties.}
  \label{tab:lora-vs-ppo}
\end{table}

\section{Pre-PPO Performance} \label{apx:pre-ppo}

We include here the results of the SFT, RM, and Hydra-SFT models on their respective datasets. Notably, the Hydra-SFT RM head usually outperforms the standard RM.

\begin{table}[H]
  \centering
  \begin{tabular}{ccc}
    \toprule
    Model & Causal PPL & Reward Accuracy   \\
    \midrule
    \multicolumn{3}{c}{GPT-4-LLM} \\
    \midrule
    OPT 1.3b Separate & 1.71  &   92.89\%   \\
    OPT 1.3b Hydra &   1.66 &   92.48\%  \\
    Llama 7b Separate &  1.48  &   93.50\%  \\
    Llama 7b Hydra &   1.47 &   95.37\% \\
    \midrule
    \multicolumn{3}{c}{Open-Assistant} \\
    \midrule
    OPT 1.3b Separate &  2.00  &   77.79\% \\
    OPT 1.3b Hydra &  1.72 &   80.97\% \\
    Llama 7b Separate & 1.33   &   76.75\% \\
    Llama 7b Hydra &  1.33 &   85.51\% \\
    \midrule
    \multicolumn{3}{c}{Learning to Summarize} \\
    \midrule
    Llama 7b Separate & 2.69  &   69.31\% \\
    Llama 7b Hydra &   2.69 &   69.36\% \\
    \midrule
    \multicolumn{3}{c}{StackExchange} \\
    \midrule
    Llama 7b Separate &  2.08  &   63.90\% \\
    Llama 7b Hydra &   2.10 &   64.20\% \\
    \bottomrule
  \end{tabular}
  \captionsetup{skip=5pt}
  \caption{Model Pre-PPO Performance results on their respective validation datasets.}
  \label{tab:preppo-all}
\end{table}

\end{document}